\documentclass[letterpaper, 10 pt, conference]{ieeeconf}

\IEEEoverridecommandlockouts                           
\usepackage{times}
\usepackage{graphics} 
\usepackage{mathtools}
\usepackage{graphicx}
\usepackage{tabularx}
\usepackage{caption}
\usepackage{wrapfig}
\usepackage{romannum}
\usepackage{amsmath,amssymb}
\usepackage{url}
\usepackage{bm}
\usepackage[table,xcdraw]{xcolor}
\usepackage{hyperref}
\usepackage{subfigure}
\usepackage{cite}

\usepackage{multirow}
\usepackage{booktabs}
\usepackage{pifont}
\usepackage{arydshln}
\usepackage{tabularx}

\usepackage[ruled,vlined]{algorithm2e}
\usepackage[noend]{algpseudocode}
\makeatletter
\let\OldStatex\Statex
\renewcommand{\Statex}[1][3]{%
  \setlength\@tempdima{\algorithmicindent}%
  \OldStatex\hskip\dimexpr#1\@tempdima\relax}
\renewcommand{\ALG@beginalgorithmic}{\normalsize}

\newcommand{\argmax}{\operatornamewithlimits{argmax}}
\newcommand{\argmin}{\operatornamewithlimits{argmin}}

\DeclareCaptionFont{mysize}{\fontsize{8}{9.6}\selectfont}
\title{\LARGE \bf Temporal Action Representation Learning 
\\for Tactical Resource Control and Subsequent Maneuver Generation}

\author{Anonymous Authors}
\author{Hoseong Jung, Sungil Son, Daesol Cho, Jonghae Park, Changhyun Choi, H. Jin Kim$^{*}$
\thanks{This work was partly supported by Institute of Information \& communications Technology Planning \& Evaluation (IITP) grant funded by the Korea government (MSIT) [NO.RS-2021-II211343, Artificial Intelligence Graduate School Program (Seoul National University)] and Hyundai Motor Chung Mong-Koo Foundation.}
\thanks{Hoseong Jung, Jonghae Park, Changhyun Choi, and H. Jin Kim are with Seoul National University, Republic of Korea.
Sungil Son is with Seoul National University and Life Assistant Robotics, Republic of Korea.
Daesol Cho is with the Georgia Institute of Technology, USA.}
\thanks{$^*$Corresponding author}}%

\begin{document}

\maketitle


\begin{abstract}
Autonomous robotic systems should reason about resource control and its impact on subsequent maneuvers, especially when operating with limited energy budgets or restricted sensing.
Learning-based control is effective in handling complex dynamics and represents the problem as a hybrid action space unifying discrete resource usage and continuous maneuvers.
However, prior works on hybrid action space have not sufficiently captured the causal dependencies between resource usage and maneuvers.
They have also overlooked the multi-modal nature of tactical decisions, both of which are critical in fast-evolving scenarios.
In this paper, we propose TART, a Temporal Action Representation learning framework for Tactical resource control and subsequent maneuver generation.
TART leverages contrastive learning based on a mutual information objective, designed to capture inherent temporal dependencies in resource-maneuver interactions.
These learned representations are quantized into discrete codebook entries that condition the policy, capturing recurring tactical patterns and enabling multi-modal and temporally coherent behaviors.
We evaluate TART in two domains where resource deployment is critical: (i) a maze navigation task where a limited budget of discrete actions provides enhanced mobility, and (ii) a high-fidelity air combat simulator in which an F-16 agent operates weapons and defensive systems in coordination with flight maneuvers.
Across both domains, TART consistently outperforms hybrid-action baselines, demonstrating its effectiveness in leveraging limited resources and producing context-aware subsequent maneuvers.
\end{abstract}

\section{INTRODUCTION}
Autonomous robotic systems are required to operate under limited resources (e.g., task allocation, computational budget, and battery capacity), making efficient resource utilization a critical requirement for real-world deployment~\cite{gerkey2004formal, afrin2021resource, dai2022multi, hai2025replanning, liao2025sa}.
This challenge is particularly critical in dynamic environments, where agents must make rapid decisions in response to fast-changing situations~\cite{hai2025replanning, liao2025sa}.
In this work, we consider a tactical decision-making problem by viewing resource usage as part of robotic actions, where each action incurs a finite resource cost.
Effective control over such resource-usage actions must be coordinated with the generation of subsequent maneuvers, as each decision determines not only how resources are consumed but also how their effects can be maximized through follow-up actions.

\begin{figure}[t!]
\centering
\includegraphics[width=\columnwidth]{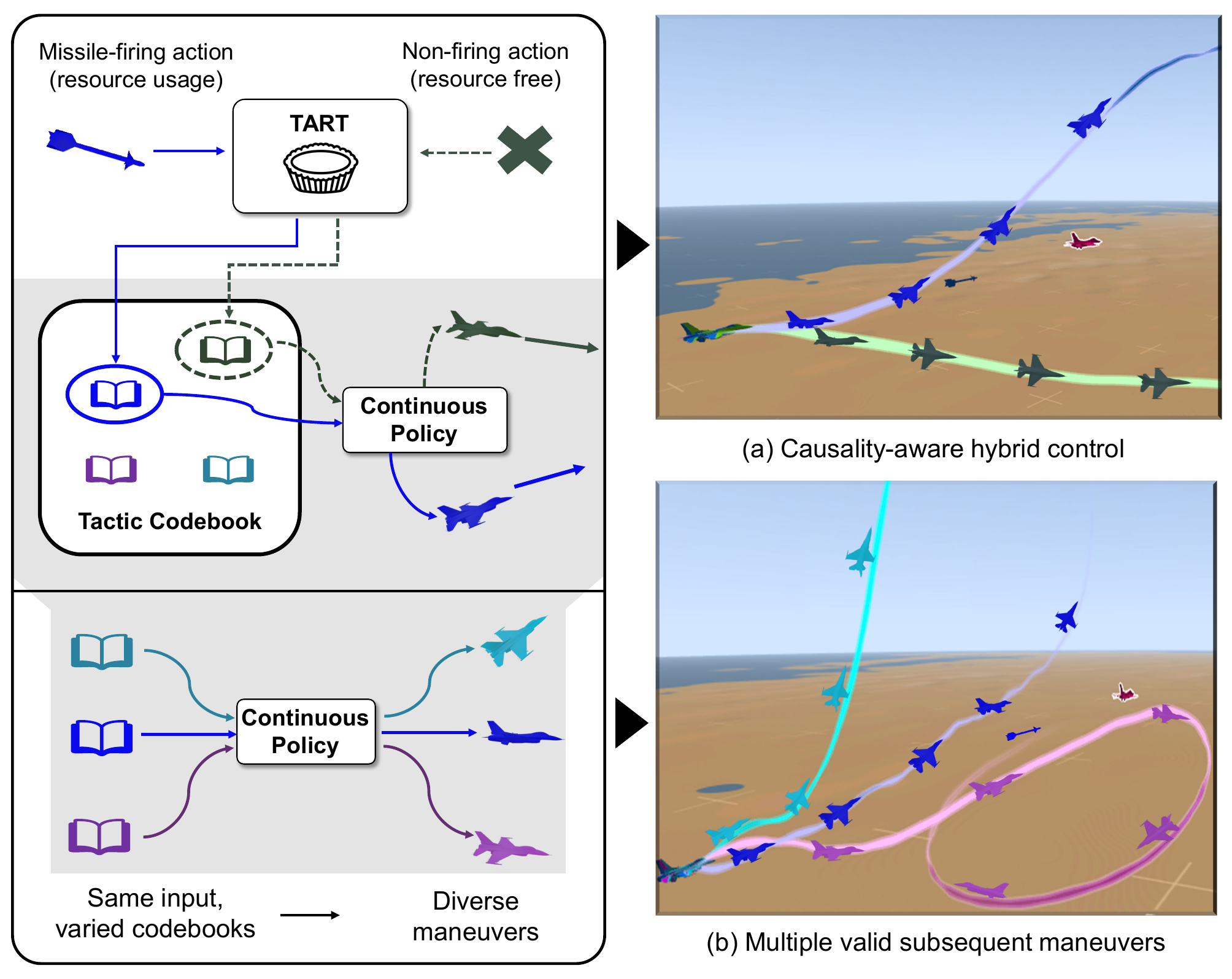}
\caption{Snapshots of tactical decision-making in an air combat scenario. (a) A discrete action (e.g., weapon release) both constrains the set of feasible follow-up maneuvers (\textit{causal dependency}) and (b) gives rise to multiple valid maneuver modes (\textit{multi-modality}). TART is designed to capture these temporal dependencies and multi-modal outcomes, conditioning the policy to select context-appropriate maneuvers.
\vspace{-20pt}
}
\label{fig:main-figure}
\end{figure}

Addressing tactical decision-making under resource constraints requires explicit modeling of hybrid action spaces that combine resource usage with maneuver control.
These hybrid structures naturally arise in practice: discrete actions govern resource usage (e.g., weapon release in Fig.~\ref{fig:main-figure}), while continuous actions guide low-level behaviors such as motion control~\cite{hausknecht2015deep}.
Although reinforcement learning (RL) has been widely applied in these domains~\cite{hausknecht2015deep, masson2016reinforcement, xiong2018parametrized, fan2019hybrid, li2021hyar}, existing approaches often overlook two aspects that are particularly critical in tactical decision-making: \textit{causal dependency} and \textit{multi-modality}.
Causal reasoning is necessary to predict how resource usage constrains future maneuvers.
Multi-modality is essential in dynamic environments such as air combat or mobile robot navigation, where the same discrete decision may lead to multiple valid follow-up maneuvers depending on the evolving situation (Fig.~\ref{fig:main-figure}).
Conventional RL approaches neglect these properties, thereby converging to a single dominant policy and limiting tactical flexibility~\cite{wu2025discrete}.

These challenges highlight the need for temporal representation learning of robotic actions that can jointly capture resource usage, causal effects, and maneuver diversity.
This necessity arises from the hierarchical structure of decision-making, where discrete resource-allocation decisions are temporally coupled with continuous maneuver controls.
Representation learning provides a means to abstract these interactions into structured latent embeddings, enabling policies to reason over temporal dependencies~\cite{ze2023visual, biza2025robot, bu2025univla} while preserving the diversity of valid maneuver outcomes~\cite{mehta2022learning ,luo2023action, zhu2023imitation}.
Such compact representations are largely absent in prior RL approaches to scalable and resource-aware robotic decision-making, underscoring the importance of temporally grounded action representations.

In this paper, we propose \textbf{T}emporal \textbf{A}ction \textbf{R}epresentation learning for \textbf{T}actical resource control and subsequent maneuver generation (TART), a framework for learning temporally grounded representations for hybrid action policies.
The core idea is to model the conditional distribution of continuous maneuvers given recent state history with the current discrete resource usage decision.
We realize this by maximizing a mutual information objective via a trajectory-level contrastive loss that aligns matched context--future pairs and separates mismatched pairs~\cite{oord2018representation}.
The resulting context embeddings are vector-quantized into a compact, interpretable codebook of tactical modes~\cite{van2017neural}.
These codes condition a factorized hybrid policy where discrete actions feed into the representation to select a tactical mode, and the resulting mode guides a continuous actor to produce a multi-modal maneuver distribution.
Our representation learning framework integrates into the standard on-policy RL loop, being updated jointly with policy optimization~\cite{schulman2017proximal}.
By grounding action representations temporally and structuring them into discrete tactical modes, TART preserves maneuver multi-modality while enforcing resource-aware consistency under hybrid action spaces.

We evaluate TART in two domains that combine resource constraints with hybrid action spaces.
The first is a maze navigation task where agents have a limited budget of discrete boost and wall penetration actions that enhance mobility alongside continuous movement.
The second is a high-fidelity air-to-air combat simulator where an F-16 agent coordinates continuous flight control with discrete weapon and defense system deployment.
Across both domains, TART outperforms standard hybrid-action baselines and generates behaviors that capture causal dependencies and support multi-modal tactical decisions.
These results underscore the importance of structured action representations that encode temporal dependencies, enabling resource-aware robotic decision-making.

The contributions of our work are as follows:

\begin{itemize}
    \item We introduce TART, a temporal action representation learning framework that unifies resource control and maneuver generation to address tactical decision-making under resource limits in hybrid action spaces.
    \item We propose a mutual information-based objective with trajectory-level contrastive learning and a vector-quantized codebook, resulting in representations that capture causal dependencies and support multi-modal, temporally coherent behaviors.
    \item We conduct extensive experiments in both maze navigation and air-to-air combat environments, demonstrating (i) superior performance over hybrid-action baselines, (ii) improved modeling of maneuver multi-modality and temporal coherence in hybrid action sequences.
\end{itemize}

\section{Background and Related Work}
\subsection{Parameterized Action Markov Decision Process}
\begin{figure*}[t!]
\centering
\includegraphics[width=1\textwidth]{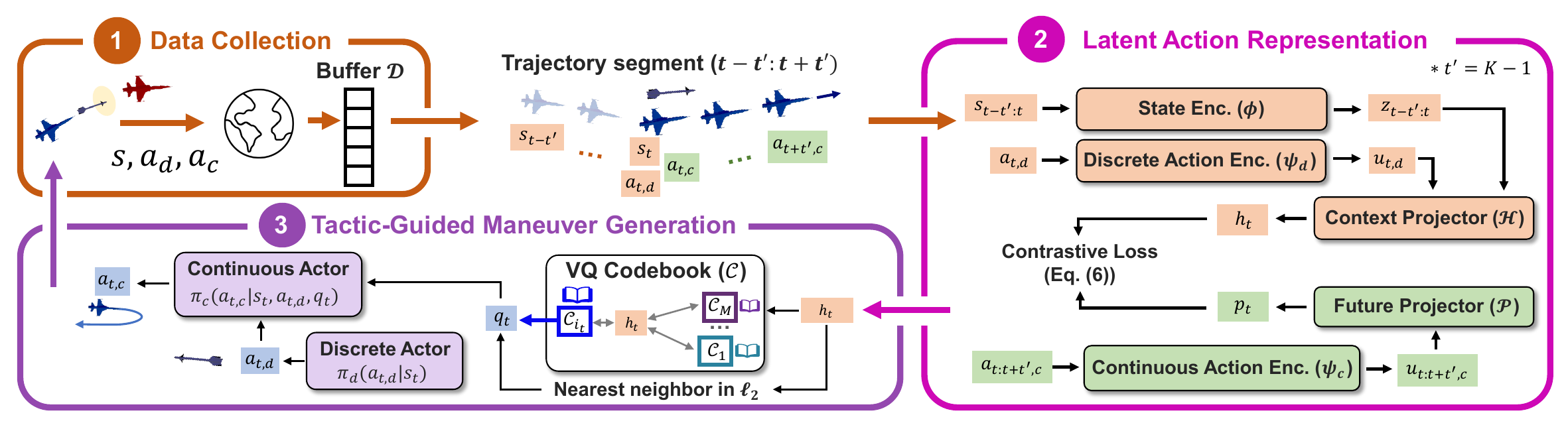}
\caption{Overview of TART: (1) The agent interacts with the environment and collects a set of trajectories. (2) A mutual information objective guides the clustering of given trajectories into multiple tactical modes through contrastive learning (Sec. III-B). (3) The resulting distinct modes are then mapped to discrete vectors via vector quantization (VQ). The continuous actor distinguishes between the modes and generates multi-modal maneuver distributions accordingly (Sec. III-C).
}
\vspace{-10pt}
\label{fig:TART}
\end{figure*}

In this paper, we build on a Parameterized Action Markov Decision Process (PAMDP) $\langle\mathcal{S}, \mathcal{X}, P, \gamma, \mathcal{R}\rangle$, defined with a state space $\mathcal{S}$, parameterized action space $\mathcal{X}$, transition function $P : \mathcal{S} \times \mathcal{X} \times \mathcal{S} \rightarrow \mathbb{R}$, discount factor $\gamma \in [0, 1)$, and reward function~$\mathcal{R} : \mathcal{S} \times \mathcal{X} \rightarrow \mathbb{R}$~\cite{masson2016reinforcement}.
Specifically, we extend the standard PAMDP framework to incorporate a discrete-continuous hybrid action space defined as Cartesian product $\mathcal{X}=\mathcal{A}_d \times \mathcal{A}_c$, where $\mathcal{A}_d = \{ a_{d,1}, ..., a_{d, m}\}$ is a finite set of $m$ discrete resource decisions and $\mathcal{A}_c \subseteq\mathbb{R}^n$ is the $n$-dimensional continuous control space.
A hybrid action is denoted $a=(a_d, a_c)$.
The agent maximizes the expected discounted return $J(\pi)=\mathbb{E}_\pi[\sum_{t=0}^\infty\gamma^t\mathcal{R}(s_t, a_t)]$, where the policy factorizes as $\pi(a_d, a_c|s)=\pi_d(a_d|s)\pi_c(a_c|s, a_d)$.
We recall the value functions, $V^\pi(s):=\mathbb{E}_\pi[\sum_{t\geq0}\gamma^t\mathcal{R}(s_t, a_t) | s_0=s]$ and $Q^\pi(s, a):=\mathcal{R}(s,a) + \gamma\mathbb{E}_{s'\sim P(\cdot|s,a)}[V^\pi(s')]$.
The optimal value functions are denoted as $V^*(s)=\sup_\pi V^\pi(s)$ and $Q^*(s,a)=\sup_\pi Q^\pi(s,a)$.
\subsection{Reinforcement Learning for Hybrid Action Spaces}
Hybrid action spaces commonly arise in practical domains such as traffic signal control~\cite{luo2024reinforcement}, financial trading~\cite{pan2022learn}, robotics~\cite{neunert2020continuous, fu2024multiagent}, where agents must coordinate discrete and continuous actions.
This formulation is also natural for resource-control problems, where discrete resource-usage decisions must be integrated with continuous maneuvering.
These settings have motivated RL methods tailored to hybrid actions, including extensions of Q-learning~\cite{masson2016reinforcement, xiong2018parametrized} and actor-critic approaches~\cite{ hausknecht2015deep, fan2019hybrid}.
More recently, Li~\textit{et~al}. proposed HyAR~\cite{li2021hyar}, which uses a conditional Variational Autoencoder (cVAE) to embed single-step hybrid actions and reduce redundancy in the enlarged action space.
However, single-step embeddings cannot capture long-horizon tactical dependencies, motivating temporally grounded action representations that couple discrete resource control with continuous maneuver generation under resource limits.
\subsection{Representation Learning for Reinforcement Learning}
Representation learning in RL has primarily focused on learning compact state representations for sample efficiency, often through self-supervised objectives such as contrastive learning~\cite{allshire2021laser, biza2025robot, oord2018representation}.
Beyond state encoding, several methods learn embeddings of state-action pairs~\cite{allshire2021laser, zheng2023taco}.
For example, TACO~\cite{zheng2023taco} maximizes mutual information between state-action pairs and future states to obtain action embeddings.
Another key problem is capturing behavioral multi-modality, where the same context can lead to diverse plausible futures~\cite{zhu2023imitation}.
To address this, another line of work discretizes trajectories into vector-quantized codebooks, encouraging multi-modal behaviors~\cite{luo2023action, van2017neural}.
However, these approaches do not capture the temporal coupling between a discrete resource-usage decision and the distribution over subsequent continuous maneuvers in hybrid action spaces.

\section{METHODS}
This section mainly introduces TART, a framework for learning discrete-continuous (hybrid) tactical representations that condition a policy to produce multi-modal hybrid actions.
TART learns via mutual information-based objectives with a practical contrastive loss to align temporal state-action representations with future maneuver sequences, enabling effective use of limited resources.
The disentangled representations are transformed into discrete, interpretable codebooks, which condition a policy network to generate multi-modal maneuver distributions.
Fig.~\ref{fig:TART} provides an overview of the proposed method.
\subsection{Objective for Temporal Action Representation Learning}
We begin by presenting the temporal action representation learning objective of TART.
The guiding principle of our method is to learn state and action representations that capture temporal dependencies essential for effective control under hybrid actions and resource constraints.
We quantify statistical dependence using mutual information (MI), denoted $\mathcal{I}(X;Y)$, which is a reparameterization-invariant measure of dependency:
\begin{equation}
    \mathcal{I}(X;Y)=\mathbb{E}_{p(x,y)} \log [\frac{p(x,y)}{p(x)p(y)}] = H(X) - H(X|Y),
    \label{eq:MI}
\end{equation}
where $X$ and $Y$ represent either raw samples or stochastic representations from a data distribution.

We define the state embedding $z_t=\phi(s_t)$ and the hybrid action embeddings $u_{t,d}=\psi_d(a_{t,d})$ and $u_{t,c}=\psi_c(a_{t,c})$ for the discrete and continuous components, where $s_t$, $a_{t, d}$, and $a_{t,c}$ denote the raw state and hybrid action components.
Here, $\phi$ and $\psi_d$, $\psi_c$ serve as encoders for the state $s_t$, and the action components $a_{t,d}$ and $a_{t,c}$, respectively.
An MI objective is adopted over a fixed horizon $K$:
\begin{equation}
    \mathbb{J}_{\text{TART}} = \mathcal{I}(u_{t:t+K-1,c};[z_{t-K+1:t}, u_{t,d}]).
    \label{eq:TART_objective}
\end{equation}
Intuitively, maximizing $\mathbb{J}_{\text{TART}}$ preserves the predictive information linking the state history (including the current state) and the discrete decision to the future continuous maneuver sequence, thereby promoting temporally coherent control.

Eq.~\eqref{eq:TART_objective} aims to learn state-action representations sufficient for optimal value estimation in hybrid action spaces.
To formalize this property targeted by TART, we extend the $Q^*$-sufficiency notions introduced by~\cite{rakelly2021which} to hybrid action settings.
Intuitively, joint $Q^*$-sufficiency is a representational property: if $(\phi, \psi_d, \psi_c)$ is jointly $Q^*$-sufficient, then the optimal action-value $Q^*$ can be evaluated without loss of information in the representation space.

\newtheorem{definition}{Definition}
\begin{definition}
    (Joint $Q^*$-sufficiency for hybrid actions).\\
    Let $\phi: \mathcal{S} \rightarrow \mathcal{Z}$, $\psi_d : \mathcal{A}_d \rightarrow \mathcal{U}_d,$ and $\psi_c: \mathcal{A}_c \rightarrow \mathcal{U}_c$ be state, discrete-action, and continuous-action encoders.
    For a set of reward functions $\mathcal{R}$, the triplet $(\phi, \psi_d, \psi_c)$ is jointly $Q^*$-sufficient w.r.t. $\mathcal{R}$ if $\forall r \in \mathcal{R}, s_1, s_2 \in \mathcal{S}, a_1, a_2 \in \mathcal{X},$ 
\begin{equation*}
\begin{aligned}
    \phi(s_1)=\phi(s_2), \, \psi_d(a_{1, d})=\psi_d(a_{2,d}), \, \psi_c(a_{1, c})=\psi_c(a_{2, c}) \\ \Rightarrow Q_r^*(s_1, (a_{1, d}, a_{1, c})) = Q_r^*(s_2, (a_{2, d}, a_{2, c})).
\end{aligned}
\end{equation*}
Equivalently, there exists a measurable $\tilde{Q}$ such that $Q_r^*(s, a)=\tilde{Q}(\phi(s), (\psi_d(a_d), \psi_c(a_c)))$.
\label{def:1}
\end{definition}
For a generic objective $\mathbb{J}$, we define the maximizer set:
\begin{equation}
    \Phi_\mathbb{J} = \argmax_{(\phi', \psi'_d, \psi'_c) \in \mathcal{F} \times \mathcal{G}_d \times \mathcal{G}_c} \mathbb{J}(\phi', \psi'_d, \psi'_c),
\end{equation}
where $\mathcal{F}, \mathcal{G}_d$ and $\mathcal{G}_c$ denote the function classes for the state, discrete-action, and continuous-action encoders.
Then an objective $\mathbb{J}$ is jointly $Q^*$-sufficient with respect to $\mathcal{R}$ if every maximizer $(\phi, \psi_d, \psi_c) \in \Phi_\mathbb{J}$ satisfies Definition~\ref{def:1}.
In our method, $\mathbb{J}$ is instantiated as Eq.~\eqref{eq:TART_objective}; under standard assumptions on the PAMDP and consistent MI estimation, this objective promotes representations approaching joint $Q^*$-sufficiency for value estimation in hybrid action settings.
Consequently, there exist measurable $\tilde{Q}, \tilde{V}$ that factor through $(\phi, \psi_d, \psi_c)$, justifying their use for actor-critic training built on these representations.
\subsection{Learning a Latent Action Representation}
Since direct computation of mutual information is typically intractable, we instead optimize a tractable lower bound using InfoNCE~\cite{oord2018representation}:
\begin{equation}
    \mathcal{I}(X;Y) \geq \log(N) - \mathcal{L}_{\text{NCE}},
    \label{eq:lower_bound}
\end{equation}
where $N$ denotes the number of samples and $\mathcal{L}_{\text{NCE}}$ is the InfoNCE loss. 
To instantiate this bound for $\mathbb{J}_{\text{TART}}$, we form pairs between a context summarizing the state history and the discrete action, and a future continuous maneuver descriptor.

Let the context projector $\mathcal{H}$ and the future projector $\mathcal{P}$ map to a common embedding space:
\begin{equation}
    h_t = \mathcal{H}([z_{t-K+1:t}, u_{t, d}]), \quad p_t = \mathcal{P}(u_{t:t+K-1,c}).
    \label{eq:projector}
\end{equation}
Given an anchor representation $h_t$, the positive sample is $p_t$ and the negatives are the in-batch items $\mathbf{v} \in \mathcal{N}_t$, where $\mathcal{N}_t :=\{\mathcal{P}(u_{s:s+K-1,c}^{(n)}) | (n, s) \neq (\text{current traj}, t)\}$ denotes embeddings from other time steps or trajectories in the batch.
The InfoNCE loss encourages the anchor to be similar to the positive while dissimilar to the negatives:
\begin{equation}
    \small
    \mathcal{L}_{\text{NCE}} = -\log \frac{\exp(\text{sim}(h_t, p_t))}{\exp(\text{sim}(h_t, p_t)) + \sum_{\mathbf{v} \in \mathcal{N}_t}\exp(\text{sim}(h_t, \mathbf{v}))},
    \label{eq:NCE}
\end{equation}
where $\text{sim}(x, y)=x^\top W y/\tau$ is a learnable bilinear similarity function with parameter $W$ and temperature $\tau>0$.
This construction directly matches the objective Eq.~\eqref{eq:TART_objective}: the context $h_t$ encodes the state trajectory and the discrete action choice, while the positive $p_t$ encodes future continuous maneuvers over the horizon $K$.
Inspired by TACO~\cite{zheng2023taco}, we additionally augment positives with temporally adjacent segments from the same episode to improve sample efficiency; negatives are temporally shuffled within a batch to reduce shortcut cues.
\subsection{Tactic-Guided Maneuver Generation}
To exploit the clustered action representations obtained via contrastive learning, we aim to represent them as quantized tactical modes that guide the policy in generating diverse maneuver distributions.
We instantiate these modes with a Vector Quantization (VQ) codebook~\cite{van2017neural}.
Let $\mathcal{C}=\{c_1, ..., c_M\}$ denote the set of learnable codebook entries, where each $c_i \in \mathbb{R}^d$.
Given a state-action segment $\tau_t=\{(s_{t'}, a_{t'}), ..., (s_{t}, a_{t})\}$, with $t'=t-K+1$, we obtain embeddings $\{z_{t'}, ..., z_{t}\}$ and $\{u_{t', d}, ..., u_{t,d}\}$, $\{u_{t', c}, ..., u_{t,c}\}$ via the encoders $\phi$, $\psi_d$, $\psi_c$ (Section~III.A.).
The context encoder used in Eq.~\eqref{eq:NCE} summarizes $h_t=\mathcal{H}([z_{t':t}, u_{t, d}])$ and we quantize by applying nearest neighbor in $\ell_2$-distance:
\begin{equation}
    i_t = \argmin_m ||h_t - c_m||_2, \quad q_t=c_{i_t}.
    \label{eq:indexing}
\end{equation}
The latent code $q_t$ serves as a tactical mode and conditions the maneuver policy.
We factorize the hybrid policy as:
\begin{equation}
    \footnotesize
    \pi_\theta(a_{t,d}, a_{t,c}|s_t, q_t) = \underbrace{\pi_d(a_{t,d}|s_t)}_{\text{discrete actor}}\cdot \underbrace{\pi_c(a_{t,c}|s_t, a_{t,d}, q_t)}_{\text{continuous actor}}.
    \label{eq:hybrid_policy_factorized}
\end{equation}
At inference, the pipeline is $s_t \xrightarrow{\pi_d} a_{t,d}$, then $h_t$ is encoded and quantized to $q_t$, and finally $\pi_c(a_{t,c}|s_t, a_{t,d}, q_t)$ generates the continuous maneuver; thus the discrete actor precedes, then the continuous actor is tactic-guided.

To learn effective quantized representations, we employ the standard vector quantization objective, consisting of a reconstruction loss and a commitment loss:
\begin{equation}
    \mathcal{L}_{\text{VQ}} = ||h_t - \hat{h}_t||^2 +  ||\text{sg}[h_t] - c_{i_t}||^2 + \beta || h_t - \text{sg}[c_{i_t}]||^2,
    \label{eq:VQ}
\end{equation}
where $\hat{h}_t$ is the reconstruction of the embedding from the codebook, $\text{sg}[\cdot]$ denotes the stop-gradient operator, and $\beta > 0$ is a hyperparameter that balances the commitment strength.
The VQ encoder is optimized to produce embeddings $h_t$ that closely match their assigned codebook entries, while the codebook itself adapts to reflect the encoder outputs.
\subsection{Training Protocol and Overall Objective}
\begin{algorithm}[t!]
	\small
	\textbf{Init} actor $\pi_d, \pi_c$ and critic $V_{\omega}$\\
	\textbf{Init} encoders $\phi, \psi_d, \psi_c$ and projectors $\mathcal{H}, \mathcal{P}$\\
        \textbf{Init} codebook $C=\{c_1, ..., c_M\}, c_i \sim \mathcal{N}(0, \sigma^2I_d)$\\
	Prepare replay buffer $\mathcal{D}$\\
	\Repeat( Stage 1: Warm-up){reaching maximum warm-up steps}{
        Sample batch $\mathcal{B}$ from replay buffer $\mathcal{D}$\\
		Update $\phi, \psi_d, \psi_c, \mathcal{H}, \mathcal{P}$ using Eqs.~\eqref{eq:NCE} and~\eqref{eq:VQ}\\
	}
	
	\Repeat( Stage 2: Main loop){reaching maximum total environment steps}{
		\For{t  $\leftarrow$  1  to $ T$}{
			\textcolor{blue}{// select discrete tactical modes to guide policy}\\
                observe $s_t$; $a_{t, d} \sim \pi_d(\cdot|s_t), z_t=\phi(s_t),$\\
                $u_{t,d}=\psi_d(a_{t,d}), h_t=\mathcal{H}([z_{t-K+1:t}, u_{t,d}])$\\
			$i_t=\argmin_m||h_t-c_m||_2, q_t=\text{sg}[c_{i_t}]$\\
			\textcolor{blue}{// generate continuous maneuvers}\\
			$a_{t,c}\sim \pi_c(\cdot|s_t, a_{t,d}, q_t)$\\ 	Execute $a_{t,d}, a_{t,c}$, observe $r_t$ and new state $s_{t+1}$\\
			Store $\{s, a_{t,d}, a_{t,c}, r, s_{t+1}, q_t, \log\pi_d, \log\pi_c\}$ in $\mathcal{D}$\\
			Sample a mini-batch of $N$ experiences from $\mathcal{D}$\\
			Update $\pi_d, \pi_c, V_\omega$ with the PPO objective~\cite{schulman2017proximal}\\
		}
		
		\Repeat{reaching maximum representation training steps}{
			Update $\phi, \psi_d, \psi_c, \mathcal{H}, \mathcal{P}$ using Eqs.~\eqref{eq:NCE} and~\eqref{eq:VQ}
		}
        }
	\caption{TART-PPO}
	\label{alg:TART-PPO}
\end{algorithm}

We train TART with a unified on-policy workflow that interleaves representation learning and policy optimization, as shown in Algorithm~\ref{alg:TART-PPO}. 
The procedure consists of two stages.
In the warm-up stage, we collect exploratory rollouts and fit the representation modules by minimizing the loss function in Eqs.~\eqref{eq:NCE} and~\eqref{eq:VQ}.
This yields latent codes that capture tactical structure before RL training.

In the main loop, PPO~\cite{schulman2017proximal} optimizes the factorized hybrid policy.
We implement the critic $V_\omega$ with a shared state backbone and two value heads.
The discrete head depends on the state $s_t$ and provides a baseline for the discrete actor $\pi_d(a_{t,d}|s_t)$. The continuous head takes the state $s_t$, the chosen discrete action $a_d$, and the tactical code $q_t$ to support the continuous actor $\pi_c(a_{t,c}|s_t, a_{t,d}, q_t)$.
The shared critic backbone improves sample efficiency, while the distinct heads provide actor-specific baselines that reduce variance, stabilize training, and capture multi-modality.

The overall loss function is as follows:
\begin{equation}
    \mathcal{L}_{\text{total}}=\underbrace{\mathcal{L}_{\text{RL}}}_{\text{PPO for } \pi_d, \pi_c} + \lambda_{\text{NCE}}\mathcal{L}_{\text{NCE}}+\lambda_{\text{VQ}}\mathcal{L}_{\text{VQ}},
    \label{eq:overall_loss}
\end{equation}
where $\mathcal{L}_{\text{NCE}}$ and $\mathcal{L}_{\text{VQ}}$ are the InfoNCE loss and VQ loss as in Eqs.~\eqref{eq:NCE}) and~\eqref{eq:VQ}. 
$\lambda_{\text{NCE}}$ and $\lambda_{\text{VQ}}$ are balancing parameters.
$\mathcal{L}_{\text{RL}}$ is the standard PPO objective with a shared critic, entropy regularization, and separate discrete and continuous heads; gradients from $\mathcal{L}_{\text{RL}}$ do not update the codebook.

\section{Evaluation Environments}
In this section, we introduce two budgeted hybrid-action environments to evaluate TART: (A) a maze navigation task where the agent can utilize limited resources to enhance mobility, and (B) a high-fidelity air combat environment with weapon and defense systems.
The overview of each environment is shown in Fig.~\ref{fig:environment}.
\subsection{Maze Navigation}
\subsubsection{Task \& Actions}
We extend POGEMA~\cite{skrynnik2024pogema} to a budgeted hybrid-action maze navigation task with a single start-goal pair and horizon $T=100$.
At each timestep $t$, the policy outputs a continuous heading vector $a_c=(x_t, y_t)\in[-1, 1]^2$, which is discretized into the nearest cardinal direction.
The environment executes one grid move accordingly.
A discrete option $a_d \in \{\textsc{NoOp}, \textsc{Penetration}, \textsc{Boost}\}$ is available with two uses per episode, and each activation lasts two steps.
\textsc{Boost} performs two sequential sub-moves along the chosen heading, and we check for collisions after each sub-move.
\textsc{Penetration} allows the agent to traverse at most one wall cell per sub-move.
If \textsc{Penetration} expires inside a wall, the agent is relocated to the nearest free cell.

\subsubsection{States \& Rewards}
We adopt the Learn-to-Follow~\cite{skrynnik2024learn} settings to reduce exploration complexity. 
In this setup, the policy receives a waypoint sequence computed by a heuristic path decider as an auxiliary input.
The state consists of the local observation, active waypoint, and remaining option budget.
The local observation is a two-channel egocentric $m \times m$ tensor centered on the agent, where $m$ denotes the observation range.
One channel encodes walls and the waypoints, while the other encodes background agent positions.
The reward combines sparse goal-reaching terms and dense shaping with a constant timestep penalty, waypoint rewards, collision penalties, and costs for option activations.

\subsubsection{Scenarios \& Metrics}
\begin{figure}[t!]
\centering
\includegraphics[width=\columnwidth]{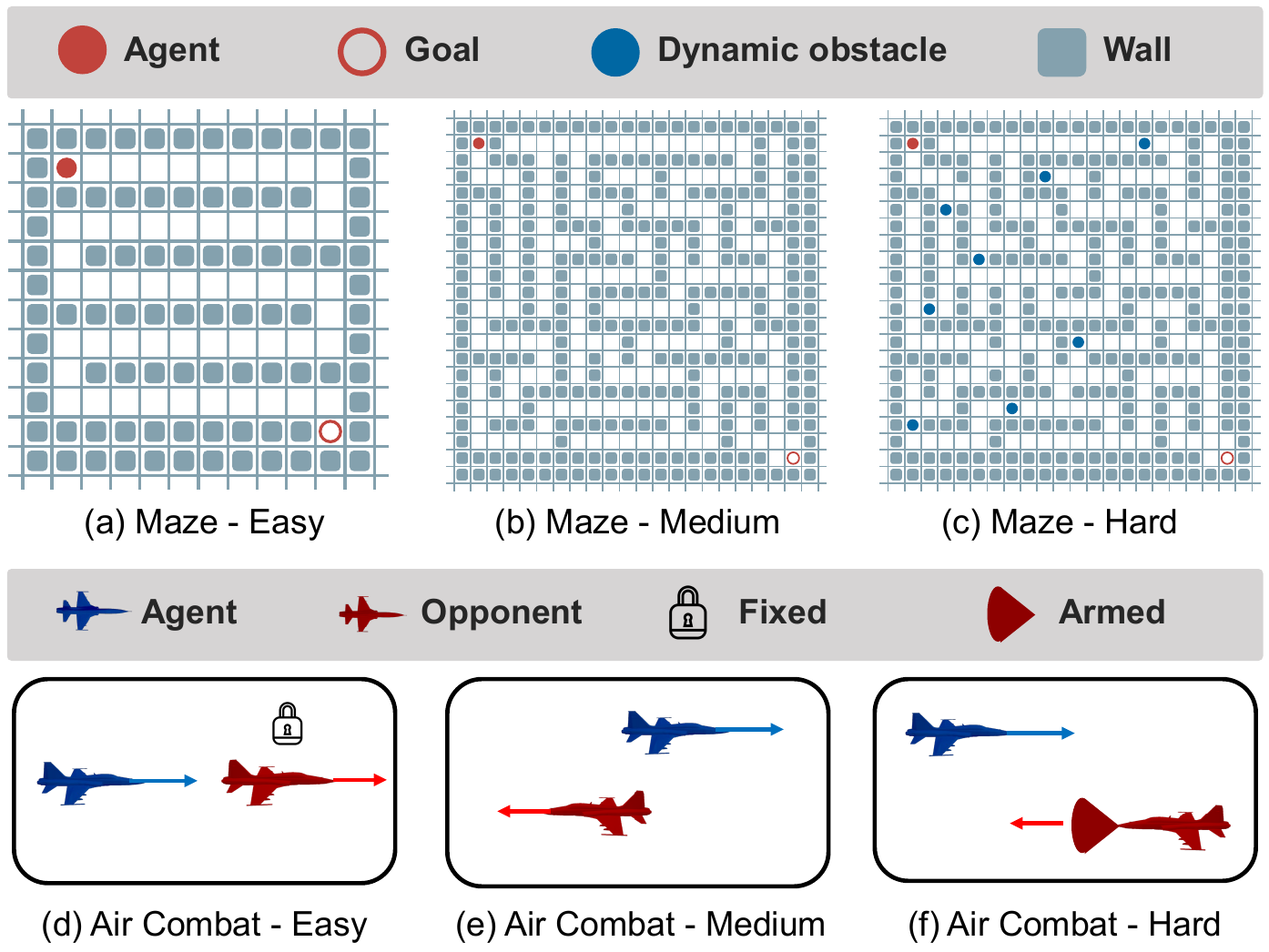}
\caption{Overviews and difficulty settings of the evaluation environments. 
(a)–(c) Maze Navigation: \emph{Easy} (trivially solvable), \emph{Medium} (complex), \emph{Hard} (dynamic obstacles). 
(d)–(f) Air-to-Air Combat: \emph{Easy} (fixed-maneuver opponent), \emph{Medium} (unarmed evasive opponent), \emph{Hard} (armed pursuing opponent).
}
\vspace{-10pt}
\label{fig:environment}
\end{figure}

Evaluation is conducted on three maze scenarios (Fig.~\ref{fig:environment}(a)-(c)) with increasing difficulty.

\begin{itemize}
    \item \emph{Easy}: $10 \times 10$ simple mazes with a trivial shortest path.
    \item \emph{Medium}: $20 \times 20$ mazes with complex layouts.
    \item \emph{Hard}: $20 \times 20$ non-stationary mazes with background agents that navigate with A* toward randomly sampled goals, which creates congestion and blockages around narrow passages.
\end{itemize}
Each difficulty includes ten distinct maze layouts to promote generalization.
We report \emph{Success Rate}, \emph{Time-to-Goal} (\textbf{TTG}) in decision steps with timeouts set to the episode horizon $T=100$, and \emph{Occupancy Coverage}, measured as the fraction of unique free cells visited to reflect path efficiency beyond goal completion.

\subsubsection{Training Details}
We follow the training details and network architecture of Learn-to-Follow~\cite{skrynnik2024learn}.
The agent uses a ResNet spatial encoder and MLP heads for the policy and critic, comprising approximately 5M parameters.
Action masking prevents invalid moves and option activations that violate budget or duration.
For each difficulty, the agent is trained jointly on ten maps and evaluated on unseen maps to assess generalization. 
The final policy is trained for 20 million steps, and evaluation uses the best checkpoint.

\subsection{Air-to-Air Combat}
\begin{table*}[t]
\vspace{-0.em}
\centering
\captionsetup{justification=centering}
 \caption{Performance (\emph{Success Rate}) for different PAMDP methods over resource-limited environments. \\ Results are averaged over five random seeds, $\pm$ indicates the standard deviation across seeds, and bold entries denote the best performance.}
 \label{tab:main_result}
\resizebox{1.0\textwidth}{!}{%
\begin{tabular}{c c >{\columncolor{gray!25}}c c c c c c c}
\toprule
 \multicolumn{2}{c}{\textbf{Task/ Scenario}} & \textbf{TART} & \textbf{PADDPG}~\cite{hausknecht2015deep} & \textbf{PDQN}~\cite{xiong2018parametrized} & \textbf{HPPO}~\cite{fan2019hybrid} & \textbf{HyAR}~\cite{li2021hyar} & \textbf{TART w/o $\mathcal{L}_{\text{NCE}}$} & \textbf{TART w/o $\mathcal{L}_{\text{VQ}}$} \\
\midrule
\multirow{2}{*}{Maze Navigation}  & Easy & $\mathbf{97.2\pm 1.3}$ & $88.2\pm 6.9$ & $85.8\pm 6.1$ & $87.2\pm 4.6$ & $94.5\pm 3.2$ & $92.8\pm 3.3$ & $95.4 \pm 1.9$ \\
                             & Medium & $\mathbf{90.8\pm 4.2}$ & $74.2 \pm 6.1$ & $68.8 \pm 6.2$ & $74.6 \pm 9.0$ & $80.6\pm 8.5$ & $89.0\pm 3.5$ & $88.2\pm 5.3$ \\ 
                             & Hard & $\mathbf{72.8\pm 9.9}$ & $38.8 \pm 10.5$ & $38.4\pm 13.5$ & $46.2\pm 6.7$ & $60.4\pm 9.1$ & $69.2\pm 5.7$ & $67.0\pm 3.2$ \\  
\midrule

\multirow{2}{*}{Air-to-Air Combat} & Easy & $\mathbf{94.8\pm 3.1}$ & $79.6\pm 5.6$ & $81.2\pm 5.9$ & $87.8\pm 4.6$ & $86.6\pm 4.8$ & $92.4\pm 5.1$ & $91.8\pm 2.2$ \\ 
                             & Medium & $\mathbf{90.8\pm 4.2}$ & $68.6\pm 5.9$ & $74.2\pm 4.3$ & $76.2\pm 4.5$ & $77.6\pm 6.3$ & $87.0\pm 6.2$ & $89.8\pm 2.6$ \\ 
                             & Hard & $\mathbf{76.8\pm 5.0}$ & $61.8\pm 7.8$ & $57.2\pm 7.4$ & $65.4\pm 3.0$ & $64.4\pm 7.2$ & $73.6\pm 6.2$ & $70.8 \pm 2.1$ \\ 
\midrule
\multicolumn{2}{r}{$\textbf{Average Success Rate}$} & $\mathbf{87.2}$ & $68.5$  & $67.6$ & $72.9$ & $77.4$ & $84.0$ & $83.8$ \\
\bottomrule
\end{tabular}%
 }
 \vspace{-0.em}
 \end{table*}

\begin{figure*}[hbt!]
\centering
\includegraphics[width=\textwidth]{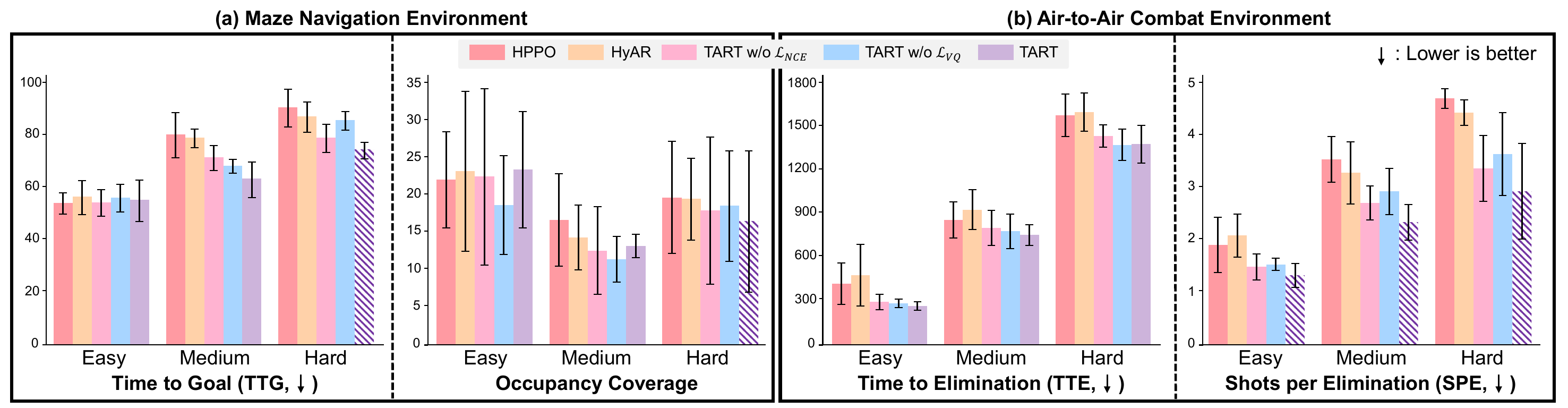}
\vspace{-10pt}
\caption{Experimental results across the designed environments and metrics. Values are averaged over five seeds and black bars indicate standard deviation. For failed episodes, TTG and TTE are set to their maximum values (100 and 1800, respectively), while SPE is reported only for successful episodes. Hatched bars indicate the results discussed in the text and exhibiting superior performance.
}
\label{fig:quan_result}
\vspace{-10pt}
\end{figure*}

\subsubsection{Task \& Actions}
We build a high-fidelity air-to-air combat environment on top of Light Aircraft Game~\cite{liu2022light} and NeuralPlane~\cite{xue2024neuralplane}.
The platform models an F-16 aircraft with six-degree-of-freedom aerodynamics in JSBSim~\cite{berndt2004jsbsim}, equipped with short-range missiles, a gun, and defensive countermeasures.
Defensive countermeasures can neutralize an incoming missile within their effective envelope.
The objective is to eliminate the opponent before being defeated.

At each timestep $t$, the agent selects a discrete option $a_d \in \{\text{\textsc{NoOp}}, \text{\textsc{Missile}}, \text{\textsc{Gun}}, \text{\textsc{Defense}}\}$ and continuous controls $a_c \in [0, 1]^4$ corresponding to aileron, elevator, rudder and throttle.
Each episode provides a budget of five missile and five defensive countermeasures, while the gun is unrestricted but constrained by strict firing conditions~\cite{pope2022hierarchical}.
Missile launch requires a valid lock-on defined by range and aspect~\cite{liu2022light}, and is effective only if the lock persists after firing.
Gun usage follows the engagement envelope of~\cite{pope2022hierarchical}, and defensive countermeasures are effective only within a circular envelope defined as twice the missile radius.
Invalid action requests are masked without consuming the budget.
The simulator runs at 10 Hz with episodes last at most three minutes, giving a maximum horizon of $T=1800$ steps.
Episodes terminate upon opponent elimination, ownship loss (including crash), or timeout.

\subsubsection{States \& Rewards}
Following prior work~\cite{bae2023deep}, the state includes ownship kinematics (attitude, altitude, velocity, acceleration) and ego-opponent geometry (range, relative aspect), extended with remaining resources for missiles and defensive countermeasures.
The reward function consists of a sparse term for opponent elimination with dense shaping.
Tail-chase shaping~\cite{bae2023deep} rewards maintaining favorable pursuit geometry behind the opponent.
Penalties apply for being shot down or crashing at low altitude, and for each missile or countermeasure used to encourage efficiency.

\subsubsection{Scenarios \& Metrics}
We evaluate agents in three air-to-air combat scenarios (Fig.~\ref{fig:environment}(d)-(f)) of increasing difficulty, defined by opponent capability and initial geometry.
\begin{itemize}
    \item \emph{Easy}: Scripted opponent~\cite{liu2022light} with simple pursuit and evade behavior and no weapons or countermeasures. Episodes start in a pursuit geometry with the ownship behind the opponent.
    \item \emph{Medium}: Scripted opponent~\cite{liu2022light} equipped with reactive weapons and countermeasures. Episodes start in neutral geometry with opposite headings, leading to a merge after roughly one turn.
    \item \emph{Hard}: A single-agent variant of the learned pursuer~\cite{he2023autonomous} equipped with reactive weapons and countermeasures. Episodes start in neutral with an immediate merge.
\end{itemize}
Evaluation metrics are \emph{Success Rate}, \emph{Time-to-Elimination} (\textbf{TTE}) measured in decision steps with a timeout $T=1800$, and \emph{Shots-per-Elimination} (\textbf{SPE}) defined as the number of missiles required to eliminate the opponent.
\emph{Success Rate} counts episodes where the agent eliminates the opponent.
For defeats, \textbf{TTE} is set to the timeout and \textbf{SPE} to the maximum budget of five.
If an elimination occurs by gun after all missile attempts are invalidated, \textbf{SPE} is also recorded as five.

\subsubsection{Training Details}
We adopt curriculum learning~\cite{bae2023deep}, which gradually increases task difficulty to stabilize and accelerate training.
The curriculum starts with pursuit geometry and no weapons, then gradually transitions to neutral geometry while enabling weapons and defensive countermeasures.
The policy and critic follow the GRU-MLP architecture of~\cite{bae2023deep}, trained with PPO~\cite{schulman2017proximal}.
Opponents are pre-trained and remain fixed during both training and evaluation.
To ensure fairness, all baselines are trained under the same curriculum.
\section{Experimental Results}
In this section, we evaluate TART's ability to learn temporally coherent action representations that follow the hybrid action structure and capture maneuver multi-modality.
We compare TART against standard PAMDP baseline methods: PADDPG~\cite{hausknecht2015deep}, PDQN~\cite{xiong2018parametrized}, HPPO~\cite{fan2019hybrid}, and HyAR~\cite{li2021hyar}.
Table~\ref{tab:main_result} reports success rates for all baselines.
Fig.~\ref{fig:quan_result} focuses on the two strongest PAMDP methods in our re-implementation, HPPO and HyAR, which also align with TART's multi-head actor design for hybrid actions.
All baselines share identical training settings for fair comparison.
\subsection{Comparison with PAMDP Baselines}
\begin{figure*}[t!]
\centering
\includegraphics[width=\textwidth]{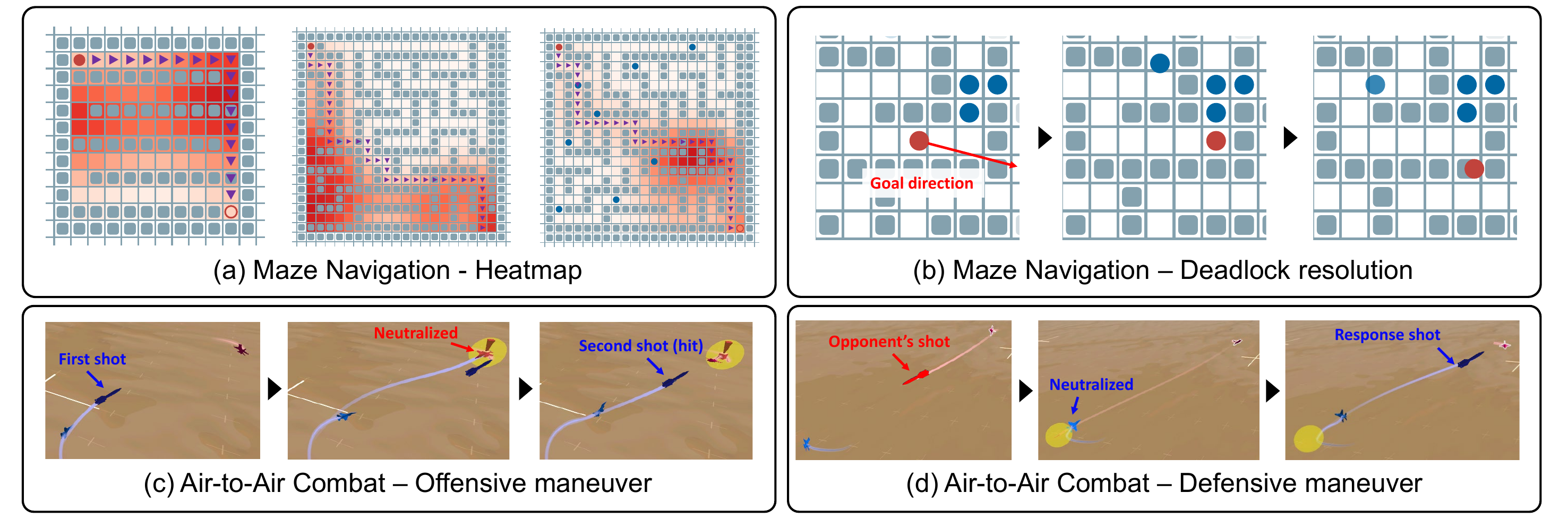}
\caption{Representative qualitative results in (a)-(b) Maze Navigation and (c)-(d) Air-to-Air Combat. (a) Heatmaps in \emph{Easy}, \emph{Medium}, and \emph{Hard} scenarios, where the triangle marks the optimal path. Red indicates higher, while white indicates lower visitation frequency. (b) Deadlock examples, where the agent employs the \textsc{Penetration} action to navigate through cluttered corridors. (c) An offensive maneuver: the agent consecutively launches shots using the \textsc{Missile} action. (d) A defensive maneuver: the agent neutralizes the opponent's missile with a \textsc{Defense} action and responds with a counter shot.
\vspace{-10pt}
}
\label{fig:qual_result}
\end{figure*}

Table~\ref{tab:main_result} shows that TART outperforms the strongest PAMDP baseline in every scenario, with \emph{Success Rate} gains from $+2.7$ to $+13.2$ points.
Across all six scenarios, TART attains an average success rate of $87.2\%$, while the best baseline method (HyAR) achieves $77.4\%$.
Fig.~\ref{fig:quan_result} further shows that these gains do not come at the cost of resource efficiency.
In Maze Navigation, TART achieves lower or comparable \textbf{TTG}, indicating more efficient path completion.
In Air-to-Air Combat, TART also demonstrates efficiency in weapon usage by yielding lower or comparable \textbf{TTE} and \textbf{SPE}.
These results imply that TART leverages the fixed action budget more effectively with coordinated continuous maneuvers, as further evidenced by the qualitative analysis in Section V-C.

\subsubsection{Maze Navigation}
Since \textit{Easy} and \textit{Medium} do not contain dynamic obstacles, all methods yield \emph{Occupancy Coverage} near the shortest-path value.
The primary difference is efficiency, where TART demonstrates lower \textbf{TTG} and higher \emph{Success Rate}, while keeping \emph{Occupancy Coverage} near the shortest-path value.
In \textit{Hard}, background agents cause corridor congestion.
TART achieves lower \emph{Occupancy Coverage} and shorter \textbf{TTG}, which suggests that it unlocks bottlenecks with $\textsc{Penetration}$ rather than exploring widely.
The PAMDP actor-head baselines (PADDPG, PDQN, HPPO) and HyAR's cVAE approach do not couple discrete actions to subsequent maneuvers.
In contrast, TART learns a temporal representation and conditions the actor on a learned code.
This design triggers $\textsc{Penetration}$ precisely when it yields progress.
This behavior is consistent with the lower \emph{Occupancy Coverage} and shorter \textbf{TTG} observed on \textit{Hard}.

\subsubsection{Air-to-Air Combat}
Across all difficulty levels, TART achieves higher \emph{Success Rate} across all difficulties than baselines while maintaining \textbf{TTE} and \textbf{SPE} similar or lower.
This outcome is critical under limited-resource constraints.
Although deploying more missiles could raise success, the non-increasing \textbf{SPE} shows that TART achieves gains without higher consumption.
The reduced \textbf{SPE} observed in \textit{Easy} and \textit{Medium} reflects TART's ability to maintain lock-on after missile releases, resulting in more valid shots.
In \textit{Hard} scenario, success depends on effective use of the remaining $\textsc{Defense}$ actions against the armed opponent.
TART employs these actions more effectively through context-aware coordination of offensive and defensive maneuvers.
Overall, these findings show that temporal coupling between resource options (\emph{Missile, Defense}) and subsequent maneuvers enables efficient resource use and effective combat performance.
\subsection{Ablation Studies}
We validate the design by ablating the contrastive loss $\mathcal{L}_{\text{NCE}}$ and the vector-quantized loss $\mathcal{L}_{\text{VQ}}$.
The contrastive loss aligns discrete resource usage decisions with the following continuous trajectories.
The vector-quantized loss induces a codebook supporting diverse maneuver patterns under similar discrete actions and state histories.
We denote the ablations as TART w/o $\mathcal{L}_{\text{NCE}}$ and TART w/o $\mathcal{L}_{\text{VQ}}$.
Removing both eliminates the learned representation and conditioning, and the method becomes close to an HPPO~\cite{fan2019hybrid} learner.

In both Maze Navigation and Air-to-Air Combat, ablating either objective reduces \emph{Success Rate}, whereas the full model consistently performs best.
In \textit{Hard} Maze Navigation, both \textbf{TTG} and \emph{Occupancy Coverage} increase in ablated models, reflecting difficulty with clutter and delayed $\textsc{Penetration}$.
In Air-to-Air Combat environment, both ablations increase \textbf{TTE} and \textbf{SPE} with identical budgets, indicating degraded launch timing and weaker follow-up maneuvers.
Variance across seeds is larger for TART w/o $\mathcal{L}_{\text{NCE}}$ than for TART w/o $\mathcal{L}_{\text{VQ}}$, suggesting that multi-modality without contrastive alignment results in unstable learning with mode drift.
These findings highlight the complementary roles of temporal alignment and maneuver diversity for mastering hybrid actions with limited resources under dynamic conditions.
\subsection{Qualitative Analysis}
\subsubsection{Maze Navigation}
Figure~\ref{fig:qual_result}(a) shows agent visitation heatmaps in Maze Navigation across different difficulty levels.
In \emph{Easy} and \emph{Medium}, the density concentrates near the shortest paths, while retaining some pre-convergence spread, indicating a reliable goal-reaching policy.
Consistent with corridor geometry, $\textsc{Boost}$ is used early to accelerate along straight segments while staying within the per-episode budget.
In \emph{Hard}, the agent can encounter deadlocks, which can be resolved by activating the $\textsc{Penetration}$ option at chokepoints.
Fig.~\ref{fig:qual_result}(b) further illustrates such deadlock cases.
Overall, the heatmap patterns align with the quantitative metrics, showing that $\textsc{Boost}$ enables faster progress along straight corridors and $\textsc{Penetration}$ resolves blockages under budget constraints.

\subsubsection{Air-to-Air Combat}
TART exhibits both strong representation ability and adaptive combat behaviors. 
In Fig.~\ref{fig:main-figure}(a), maneuvers differ with or without missile launch, showing causal dependency between discrete and continuous actions.
Figure~\ref{fig:main-figure}(b) demonstrates multi-modality, where distinct VQ codes yield different maneuvers under the same conditions.
Figures~\ref{fig:qual_result}(c)-(d) illustrate adaptive maneuvers in dynamic combat scenarios.
In (c), the agent executes an offensive maneuver with consecutive $\textsc{Missile}$ actions.
In (d), it performs a defensive maneuver, neutralizing the opponent's missile with a $\textsc{Defense}$ action and following with a counter shot.
These results show that integrating TART enables the RL agent to capture causal dependencies and multi-modality while coordinating offensive and defensive options in a context-aware manner.
\section{CONCLUSION}
This paper introduces TART, a representation learning framework for reinforcement learning in hybrid action spaces, designed to enable effective resource control and maneuver generation.
Across both maze navigation and air-to-air combat domains, TART consistently outperforms representative baselines, underscoring the importance of temporally grounded action representations for resource-constrained decision-making.
Our study is limited to simulated environments with simplified resource constraints, modeled primarily as action budgets rather than persistent physical costs.
As future work, we plan to incorporate real-world considerations such as energy consumption, communication bandwidth, and other hardware-level limitations, and extend the evaluation of TART to broader settings, including multi-agent and lifelong learning.
\addtolength{\textheight}{0cm}   

\bibliographystyle{IEEEtran}
\bibliographystyle{plainnat}


\end{document}